\documentclass[11pt,a4paper]{article}
\usepackage[hyperref]{emnlp2020}
\usepackage{times}
\usepackage{latexsym}
\usepackage{booktabs}       
\usepackage{graphicx}

\usepackage{microtype}

\aclfinalcopy 
\def\aclpaperid{094} 

\title{Big Green at WNUT 2020 Shared Task-1: Relation Extraction as Contextualized Sequence Classification}

\author{Chris Miller \\
  Department of Computer Science \\
  Dartmouth College, Hanover, NH \\
  \texttt{chris.20@dartmouth.edu} \\\And
  Soroush Vosoughi \\
  Department of Computer Science \\
  Dartmouth College, Hanover, NH \\
  \texttt{soroush.vosoughi@dartmouth.edu} \\}

\date{}

\begin{document}
\maketitle
\begin{abstract}
Relation and event extraction is an important task in natural language processing. We introduce a system which uses contextualized knowledge graph completion to classify relations and events between known entities in a noisy text environment. We report results which show that our system is able to effectively extract relations and events from a dataset of wet lab protocols.
\end{abstract}

\section{Introduction}
Wet lab protocols specify the steps and ingredients required to synthesize chemical and biological products. The majority of wet lab protocols are formatted as natural language, designed for human lab workers to interpret and carry out. Protocols are formatted differently depending on lab norms and the author writing them, and may include spelling mistakes, nonstandard abbreviations, colloquial phrasing, and assumptions that may not be obvious to readers from outside of the author's lab or field. 

Automated extraction of events, relations, and entities from this noisy language data enables standardized tracking of lab protocols, and is an important step forward for the automated reproduction of scientific results. We examine the problem of automatically identifying and classifying events and relations between entities as part of Shared Task 1 at WNUT 2020 \cite{tabassum2020wlp}. This shared task works with the Wet Lab Protocol Corpus (WLPC) introduced by \citet{kulkarni2018annotated}. The WLPC dataset consists of wet lab protocols drawn from an open-source database and annotated by a group of human annotators which included subject matter experts. 

\section{Prior Work}
Past approaches to relation and event extraction from wet lab data have included systems based on propagating information across graphs. \citet{spanrel} introduce an end-to-end system, called SpanRel, for identifying and labeling text spans and the relations between them using any text embedding model. The DyGIE and DyGIE++ systems, meanwhile, learn to propagate useful information across graphs of coreferences, relations, and events, allowing long-distance contextual information to support relation and event extraction tasks based on sliding window BERT embeddings of the text \cite{dygie,dygiepp}. 

\section{Knowledge Graphs}
\label{sec:kg}
Knowledge graphs are graph representations of the relations between entities \cite{schneider1973course}. In a typical knowledge graph construction, graph nodes represent entities, while edges of different types represent relations between entities. As an example, a knowledge graph may contain nodes for "United Kingdom" and "United Nations" with an edge of type "Member-Of" between "United Kingdom" and "United Nations." 

Because knowledge graphs are generated from imperfect information, they represent a subset of information about their component nodes and thus suffer from incompleteness. Incompleteness means that edges representing relations which exist between nodes in reality are not present in the graph (for example, if the knowledge graph contains the "United Kingdom" and "United Nations" nodes but does not contain the "Member-Of" relation between them). This property of knowledge graphs has given rise to efforts to identify missing relations between entities, a task referred to as knowledge graph completion \cite{lin2017learning}.

There are obvious parallels between knowledge graph completion and relation extraction from text given prelabeled entities; namely that both tasks require identifying a relation (if one exists) between a given pair of entities. We therefore develop a model which represents the task of extracting relations from wet lab protocols as a knowledge graph completion problem.

\section{Methodology}
\label{sec:methodology}
\begin{figure*}[h!]
    \centering
    \includegraphics[width=\textwidth]{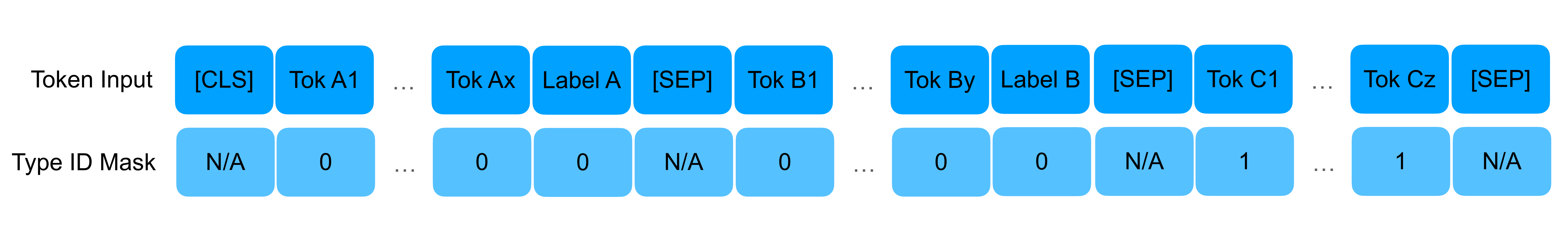}
    \caption{Input tokens and type ID masks for BERT pretraining.}
    \label{fig:tokens}
\end{figure*}

\subsection{Representing relation extraction as sequence classification}
Relation classification requires the input of two target entities to predict a relation between them. Therefore, to formulate relation extraction as relation classification, we must identify target entity pairs. A basic approach might be to simply sample each possible pair of entities in both possible orders (bidirectional sampling is required because relations are order-dependent). 

This sampling strategy, however, ignores structural information about the data. Protocols are separated into lines with one line for each step, and relations and events typically occur between entities which are close together. 

This na\"{i}ve approach also introduces computational problems. The number of possible entity pairs for $n$ entities is $n^2 - n$, which produces a high number of entity pairs as the number of entities grows. We find that real relations represent just $0.37\%$ of the possible relations in the WLPC training data, indicating that a system which enumerates all possible entity pairs would have to be exceptionally accurate to be effective. 

The structural features of the data enable us to reduce the scope of our evaluation by focusing only on entities which are close to each other. We initially evaluated based on only considering entity pairs in the same step. By analyzing the training data, we find that 99\% of true relations are between entities which contain less than 14 tokens between them. We thus restrict our analysis to entity pairs which are less than 14 tokens apart. Using this method (based on training data statistics) we are able to maintain 99\% of true relations while reducing the total number of relations evaluated by 41\% over a sentence based approach and improving our precision substantially.

\subsection{Contextualization}

One distinction between knowledge graph completion and relation extraction is important to consider. In a knowledge graph, nodes are unique and any given relation between two nodes always exists. In relation extraction from text, nodes are not unique. Consider the following protocol instruction:

    "Separate 5mL of the solution and add 5mL water to replace the removed volume."

In this protocol, the "5mL" entity of type measurement which refers to the solution is distinct from the "5mL" entity of type measurement which refers to the water. The action "Separate" acts on the former, but not on the latter, while the action "combine" acts on the latter, but not on the former. We handle this discrepancy by adding a local context sequence, identifying the targeted entities in-text. 

We generate this context sequence by taking the tokens corresponding to the $n$ sentences surrounding the target entity tokens as contextual information. We find empirically that $n=1$ provides the best performance, and that higher values of $n$ tend to cause overfitting. 

To resolve the issue of ambiguous entity reference in a sequence where multiple entities share the same text (as above), we identify entities in-context. To do this, we add entity label tokens ([EntA] and [EntB]) surrounding the referenced entities in the context, tagging them for easy identification.

\subsection{Relation Classification}
Once we have extracted a set of viable entity pair candidates, given two labeled candidate entities $E_a$ and $E_b$, and surrounding context $C$ we attempt to achieve two tasks: identifying whether or not a relationship is present between the two entities, and if there is, to classify the relationship between the entities using a knowledge graph completion approach.

Prior work has introduced the idea of using language models to formulate relation prediction between entities in a knowledge graph as a sequence classification task \cite{yao2019kg}. Pre-trained language models such as ELMo and BERT have seen widespread success when fine-tuned for use in sequence classification tasks \cite{devlin2019bert, vaswani2017attention, peters2018deep}. 

We finetune a BERT model provided by the HuggingFace library to perform relation prediction based on multi-sequence classification \cite{Wolf2019HuggingFacesTS}. We finetune for 15 epochs, using an initial learning rate of $5 \times 10^{-5}$ and an input size of 100 tokens. Hyperparameters were determined via grid search over the development set.

\subsection{Sequence Parameters}

We formulate our input sequence as shown in Figure~\ref{fig:tokens}. Token Input indicates the actual tokens which are passed into the model. For this field, [CLS] represents the start of the sequence, Tok An represents the nth token of entity A, and Tok Bn and Tok Cn represent the equivalent tokens for entity B and the context sequence, Label A and Label B are the labels for entities A and B, and [SEP] indicates separators between different sequence sources.

Type ID Mask represents the token type IDs passed to the BERT model. These binary type IDs indicate different sequence sources for multi-sequence problems such as this one. For example, when performing a classification task with two sequences, tokens from the first sequence would have a type ID of 0, and tokens from the second sequence would have a type ID of 1. The type IDs improve learning stability for BERT, ensuring that the model is able to distinguish between different sources of data.

Typically, three-sequence classification tasks in BERT are handled by masking in a 0-1-0 style (ie, the type ID mask for sequence 1 is 0, the type ID mask for sequence 2 is 1, and the type ID mask for sequence 3 is 0 again).  We find that the distinct information types of entity information and contextual information mean that labeling a sequence of Entity-Entity-Context sequences as 0-1-0 is ineffective, as BERT is not able to effectively learn the difference between context and entity information. We instead use the type-mask format 0-0-1, labeling labeled entity tokens 0 and context tokens 1. This method improves training stability and increases model performance substantially. We suggest that differences in sequence information type is the most important metric for determining type ID mask.

\section{Results \& Discussion}
\label{sec:results}

\begin{table*}[h!]
    \centering
    \caption{Results by relation type on withheld WLPC test data.}
    \begin{tabular}{lcccc}
        \toprule
        Relation Type & Precision & Recall & F1-Score & Support \\
        \midrule
        Site & 0.72 & 0.82 & 0.77 & 1622 \\
        Setting & 0.70 & 0.86 & 0.77 & 2034 \\
        Measure-Type-Link & 0.50 & 0.95 & 0.65 & 275 \\
        Coreference-Link & 0.42 & 0.53 & 0.47 & 286 \\
        Mod-Link & 0.66 & 0.88 & 0.75 & 3429 \\
        Count & 0.62 & 0.80 & 0.70 & 183 \\
        Meronym & 0.22 & 0.72 & 0.34 & 558 \\
        Using & 0.52 & 0.74 & 0.61 & 1120 \\
        Measure & 0.69 & 0.95 & 0.80 & 2370 \\
        Commands & 0.03 & 0.50 & 0.05 & 12 \\
        Of-Type & 0.56 & 0.62 & 0.58 & 193 \\
        Or & 0.22 & 0.66 & 0.33 & 193 \\
        Product & 0.13 & 0.55 & 0.21 & 42 \\
        Acts-On & 0.68 & 0.89 & 0.77 & 4072 \\
        \midrule
        Micro-Avg & 0.61 & 0.86 & 0.71 & 16354 \\
        Macro-Avg & 0.44 & 0.70 & 0.52 & 16354 \\
        \bottomrule
    \end{tabular}
    \label{tab:test_results}
\end{table*}

Our results, shown in Table~\ref{tab:test_results}, show that our system is able to effectively identify many types of relations even given this noisy data format. More specifically, this approach is able to identify relations and events with extremely high recall (as high as .95 for Measure and Measure-Type-Link relations). 

Our approach is relatively weak in precision. This is likely due to our formulation of the task as an evaluation of potential entity pairs. We find that our system classifies relations with an accuracy of 93\% on the development set, but because there are many more possible pairings between entities in a given protocol than there are actual pairings, even a system with high accuracy can incorrectly predict nonexistent relations. 

We reduce the number of possible entity pairings generated by applying a distance heuristic discussed in Section~\ref{sec:methodology}. We found that tuning the amount of entity pair candidates evaluated impacted results significantly (for example, our development set F1-score rose almost 20\% when using a token-based distance metric rather than a sentence-based distance metric for selecting candidate entity pairs). The recall of our results suggests that our distance heuristic is effective at dramatically reducing the number of evaluated entity combinations without removing too many valid combinations, but the precision indicates that it may be valuable to modify or find an alternative method for producing candidate entity pairs. This could include a method which considers contextual information, instead of focusing only on the token distance between a pair of entities. 

Class-specific result analysis allows us to identify where our system struggles. One such area is classes which occur less frequently in the data. Our macro-average F1-score is 0.69 for the seven most frequent relation classes (each of these has over 1000 examples in the training data), versus a macro-average of 0.43 for the seven least frequent relation classes (each of which has less than 1000 examples in the training data). BERT and similar language-embedding models rely on large quantities of training data, and class performance suffering due to lack of training data is not unexpected here. We expect that collection of more data for imbalanced classes could improve performance of predictions for those classes substantially.

Recent prior work has shown that BERT and other language embedding models can become overly reliant on simple patterns in the data. \citet{wnut-bert} showed that the addition of the text "10 deaths" to uninformative tweets about COVID-19 caused a BERT based system to mistakenly label them as informative. We anticipate that this effect may make our system more prone to failure in edge cases, where basic clues that the model has learned in terms of entity type patterns or contextual patterns are not present. A potential solution for this problem is to augment the training data using examples which do not have certain attributes (for example, masking entity labels). This may reduce the model's tendency to learn from basic patterns rather than true relationships between text and a relation or event class.

\section{Conclusion \& Future Work}

We show that contextualized knowledge graph completion using sequence classification can perform effectively on a relation extraction task in a noisy and specialized domain. Our model effectively identifies relations and events in the data, and our work leaves open many avenues for future work.

As discussed in Section~\ref{sec:results}, our system is sensitive to how candidate entity pairs are selected. We use a distance heuristic based on statistics of the training data to achieve our results, but we anticipate that more sophisticated methods for identifying promising candidate entity pairs could improve our results. We also suggest that our results could be improved by using a domain-specific model such as SciBERT or BioBERT (models trained on scientific papers and abstracts respectively). Prior work shows that these models often outperform standard BERT models on scientific data \cite{beltagy2019scibert, lee2020biobert}.

We believe that our results and the results of any systems which require training or fine-tuning large models would be improved by increasing available training data. Finding an effective method for augmenting existing training data and generating or collecting new training data (artificial or real) is a valuable route for further study.

Finally, we are interested in further investigation of representing relation and event identification as graph completion. Link prediction systems which support a variety of edge labels could allow us to leverage structural data from a protocol relation graph. This could enable the identification of relations which are improbable or those which may be missing from the predictions.

\bibliographystyle{acl_natbib}
\bibliography{emnlp2020}

\begin{thebibliography}{15}
\expandafter\ifx\csname natexlab\endcsname\relax\def\natexlab#1{#1}\fi

\bibitem[{Beltagy et~al.(2019)Beltagy, Lo, and Cohan}]{beltagy2019scibert}
Iz~Beltagy, Kyle Lo, and Arman Cohan. 2019.
\newblock Scibert: A pretrained language model for scientific text.
\newblock In \emph{Proceedings of the 2019 Conference on Empirical Methods in
  Natural Language Processing and the 9th International Joint Conference on
  Natural Language Processing (EMNLP-IJCNLP)}, pages 3606--3611.

\bibitem[{Chauhan(2020)}]{wnut-bert}
Kumud Chauhan. 2020.
\newblock Neu at wnut-2020 task 2: Data augmentation to tell bert that death is
  not necessarily informative.
\newblock \emph{arXiv preprint arXiv:2009.08590}.

\bibitem[{Devlin et~al.(2019)Devlin, Chang, Lee, and
  Toutanova}]{devlin2019bert}
Jacob Devlin, Ming-Wei Chang, Kenton Lee, and Kristina Toutanova. 2019.
\newblock Bert: Pre-training of deep bidirectional transformers for language
  understanding.
\newblock In \emph{HLT-NAACL}.

\bibitem[{Jiang et~al.(2019)Jiang, Xu, Araki, and Neubig}]{spanrel}
Zhengbao Jiang, Wei Xu, Jun Araki, and Graham Neubig. 2019.
\newblock Generalizing natural language analysis through span-relation
  representations.
\newblock \emph{arXiv}, pages arXiv--1911.

\bibitem[{Kulkarni et~al.(2018)Kulkarni, Xu, Ritter, and
  Machiraju}]{kulkarni2018annotated}
Chaitanya Kulkarni, Wei Xu, Alan Ritter, and Raghu Machiraju. 2018.
\newblock An annotated corpus for machine reading of instructions in wet lab
  protocols.
\newblock In \emph{HLT-NAACL}, pages 97--106.

\bibitem[{Lee et~al.(2020)Lee, Yoon, Kim, Kim, Kim, So, and
  Kang}]{lee2020biobert}
Jinhyuk Lee, Wonjin Yoon, Sungdong Kim, Donghyeon Kim, Sunkyu Kim, Chan~Ho So,
  and Jaewoo Kang. 2020.
\newblock Biobert: a pre-trained biomedical language representation model for
  biomedical text mining.
\newblock \emph{Bioinformatics}, 36(4):1234--1240.

\bibitem[{Lin et~al.(2017)Lin, Liu, Wang, Yue, and Lin}]{lin2017learning}
Hailun Lin, Yong Liu, Weiping Wang, Yinliang Yue, and Zheng Lin. 2017.
\newblock Learning entity and relation embeddings for knowledge resolution.
\newblock \emph{Procedia Computer Science}, 108:345--354.

\bibitem[{Luan et~al.(2019)Luan, Wadden, He, Shah, Ostendorf, and
  Hajishirzi}]{dygie}
Yi~Luan, Dave Wadden, Luheng He, Amy Shah, Mari Ostendorf, and Hannaneh
  Hajishirzi. 2019.
\newblock A general framework for information extraction using dynamic span
  graphs.
\newblock In \emph{HLT-NAACL}, pages 3036--3046.

\bibitem[{Peters et~al.(2018)Peters, Neumann, Iyyer, Gardner, Clark, Lee, and
  Zettlemoyer}]{peters2018deep}
Matthew~E Peters, Mark Neumann, Mohit Iyyer, Matt Gardner, Christopher Clark,
  Kenton Lee, and Luke Zettlemoyer. 2018.
\newblock Deep contextualized word representations.
\newblock \emph{arXiv preprint arXiv:1802.05365}.

\bibitem[{Schneider(1973)}]{schneider1973course}
Edward~W Schneider. 1973.
\newblock Course modularization applied: The interface system and its
  implications for sequence control and data analysis.

\bibitem[{Tabassum et~al.(2020)Tabassum, Xu, and Ritter}]{tabassum2020wlp}
Jeniya Tabassum, Wei Xu, and Alan Ritter. 2020.
\newblock {WNUT-2020 Task 1: Extracting Entities and Relations from Wet Lab
  Protocols}.
\newblock In \emph{Proceedings of EMNLP 2020 Workshop on Noisy User-generated
  Text (WNUT)}.

\bibitem[{Vaswani et~al.(2017)Vaswani, Shazeer, Parmar, Uszkoreit, Jones,
  Gomez, Kaiser, and Polosukhin}]{vaswani2017attention}
Ashish Vaswani, Noam Shazeer, Niki Parmar, Jakob Uszkoreit, Llion Jones,
  Aidan~N Gomez, {\L}ukasz Kaiser, and Illia Polosukhin. 2017.
\newblock Attention is all you need.
\newblock In \emph{Advances in Neural Information Processing Systems
  (NeurIPS)}, pages 5998--6008.

\bibitem[{Wadden et~al.(2019)Wadden, Wennberg, Luan, and Hajishirzi}]{dygiepp}
David Wadden, Ulme Wennberg, Yi~Luan, and Hannaneh Hajishirzi. 2019.
\newblock Entity, relation, and event extraction with contextualized span
  representations.
\newblock In \emph{EMNLP}, pages 5788--5793.

\bibitem[{Wolf et~al.(2019)Wolf, Debut, Sanh, Chaumond, Delangue, Moi, Cistac,
  Rault, Louf, Funtowicz, Davison, Shleifer, von Platen, Ma, Jernite, Plu, Xu,
  Scao, Gugger, Drame, Lhoest, and Rush}]{Wolf2019HuggingFacesTS}
Thomas Wolf, Lysandre Debut, Victor Sanh, Julien Chaumond, Clement Delangue,
  Anthony Moi, Pierric Cistac, Tim Rault, Rémi Louf, Morgan Funtowicz, Joe
  Davison, Sam Shleifer, Patrick von Platen, Clara Ma, Yacine Jernite, Julien
  Plu, Canwen Xu, Teven~Le Scao, Sylvain Gugger, Mariama Drame, Quentin Lhoest,
  and Alexander~M. Rush. 2019.
\newblock Huggingface's transformers: State-of-the-art natural language
  processing.
\newblock \emph{ArXiv}, abs/1910.03771.

\bibitem[{Yao et~al.(2019)Yao, Mao, and Luo}]{yao2019kg}
Liang Yao, Chengsheng Mao, and Yuan Luo. 2019.
\newblock Kg-bert: Bert for knowledge graph completion.
\newblock \emph{arXiv preprint arXiv:1909.03193}.

\end{thebibliography}

\end{document}